\documentclass{article}
\usepackage{multirow}
\usepackage{xcolor}
\usepackage{spconf,amsmath,graphicx,hyperref}
\usepackage{booktabs}
\usepackage{amssymb}

\title{PSEE: Progressive Sensor Event Expansion for Point-Supervised Temporal Action Localization}

\name{Jiaxi Yin, Ge Wang, Han Ding, Fei Wang
$^*$~\thanks{$^{*}$ Corresponding author.}
}
\address{
Xi'an Jiaotong University, Xi'an, China
\\ \textit{jiaxiyin@stu.xjtu.edu.cn, \{gewang, dinghan,feynmanw\}@xjtu.edu.cn }
}

\begin{document}

\maketitle

\begin{abstract}
Temporal action localization (TAL) in wearable sensor streams identifies
action classes and temporal boundaries, enabling finer-grained activity
understanding than conventional action recognition.
However, training typically requires costly start--end annotations
for every action instance.
To reduce this burden, we study point-supervised TAL, where each
instance is labeled with only one timestamp and its class.
We propose Progressive Sensor Event Expansion (PSEE), which combines
semantic activations, sensor-specific transition evidence, and adaptive
temporal ownership to recover point-supervised pseudo segments.
These segments supervise standard TAL detectors without modifying
their inference procedures.
Cross-subject experiments on four inertial-sensing benchmarks demonstrate
improved pseudo-boundary quality over adapted point-supervised baselines,
compatibility with different TAL detectors, and robustness to point sampling.
Code is available at \url{https://github.com/joeeeeyin/PSEE}.
\end{abstract}


\section{Introduction}
\label{sec:intro}

Temporal action localization (TAL) in wearable sensor streams identifies
action classes and temporal boundaries, enabling point-supervised activity
analysis~\cite{bock2024temporal,cui2026light}.
Applications include outdoor workout analysis
(WEAR~\cite{bock2024wear}), laboratory procedure monitoring
(WetLab~\cite{scholl2015wearables}), and daily activity and
postural-transition analysis (XRFV2~\cite{lan2025xrfv2}, RWHAR~\cite{sztyler2016body} and
SBHAR~\cite{reyesortiz2016transition}).
Despite advances in sensor TAL~\cite{bock2024temporal,cui2026light},
training typically requires precise start--end annotations, which are
labor-intensive for continuous multichannel signals with gradual
activity transitions.

Point-supervised TAL reduces annotation costs by assigning only one
timestamp and a class label to each action instance, leaving its
temporal extent unknown~\cite{ma2020sf,lee2021learning,zhang2024hr,
xia2024realigning,liu2024smbd,liu2025boosting, vahdani2024potloc, yin2023proposal}.
In video TAL, SF-Net~\cite{ma2020sf} expands sparse annotations and mines
pseudo background frames, while SMBD~\cite{liu2024smbd} searches for
boundaries between action points and background anchors.
In wearable sensing, Xia et al.~\cite{xia2024timestamp} infer dense
sample-level labels through class-activation-guided representation
learning and order-preserving optimal transport.
Their objective, however, is sample-wise segmentation rather than
point-supervised localization with explicit start--end segments.
These differences motivate a point-to-segment approach that combines
semantic evidence with sensor-specific temporal transitions to recover
point-supervised supervision.

We propose Progressive Sensor Event Expansion (PSEE)
to recover pseudo segments from sparse point annotations.
A point localizer produces class activations and
transition responses, with the latter combining network predictions
and a sensor-specific prior based on adjacent signal variations
and cross-channel change consistency.
During pseudo-boundary inference (PBI), neighboring annotations
and fused transition responses define adaptive ownership
regions that constrain boundary search for each point.
Within each region, class-activation support endpoints, transition
peaks, and ownership limits provide boundary candidates.
Rank-based selection balances action presence, class-activation
contrast, transition alignment, and support coverage to select
a pseudo segment per point.
These segments supervise standard TAL detectors, with segment-level
reliability scores reweighting only the localization loss.
The detector architecture and inference remain unchanged,
without point localizer or PBI at inference.

\begin{figure*}[t]
    \centering
    \includegraphics[width=0.8\textwidth]{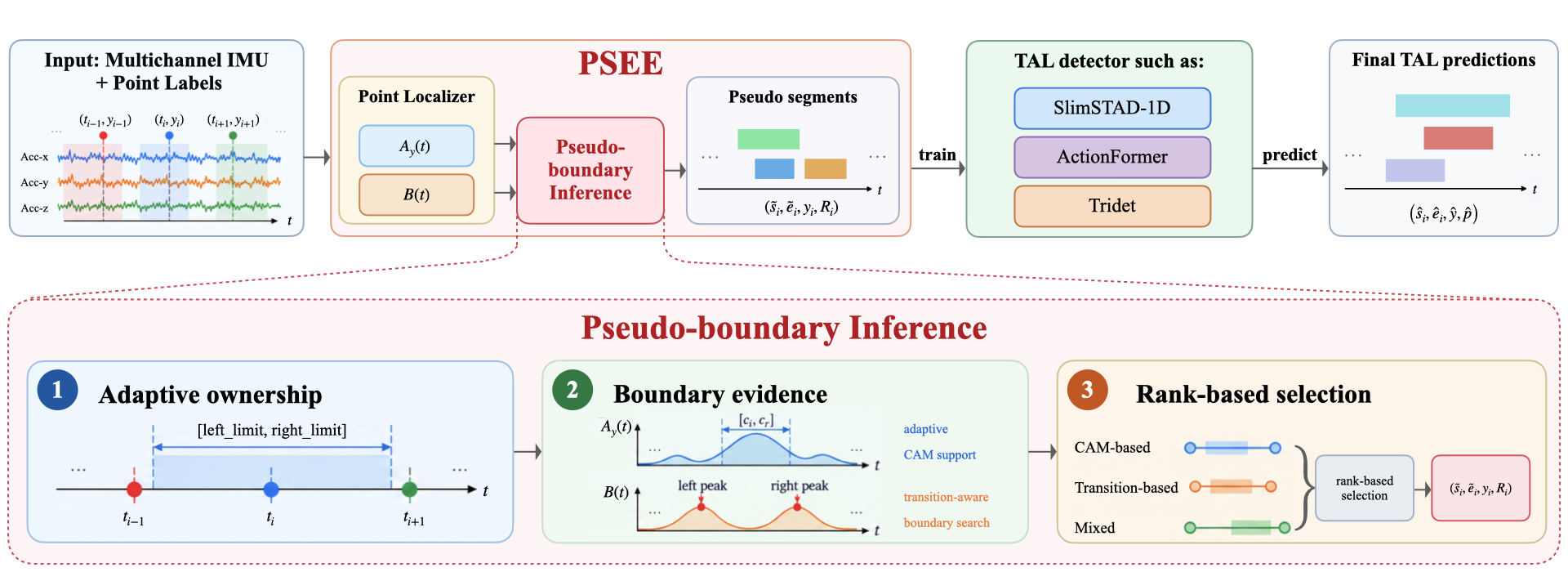}
    \caption{Overview of PSEE for point-supervised sensor TAL. Pseudo segments generated from sparse action points supervise a standard TAL detector during training.}
    \label{fig:framework}
    \vspace{-8pt}

\end{figure*}

We evaluate PSEE on the four benchmarks under a cross-subject
leave-one-subject-out (LOSO) protocol.
With SlimSTAD~\cite{cui2026light}, PSEE achieves 55.70\% mAP averaged over the four
datasets and tIoU thresholds $0.3{:}0.1{:}0.7$, compared with 35.95\%
for OPOT-Adapt~\cite{xia2024timestamp} and 34.18\% for
SFNet-Adapt~\cite{ma2020sf}.
Further experiments with ActionFormer~\cite{zhang2022actionformer}
and TriDet~\cite{shi2023tridet} confirm compatibility with different
TAL detectors. Our contributions are:
\vspace{-4pt}
\begin{enumerate}
    \setlength{\parskip}{0pt}
    \setlength{\parsep}{0pt}
    \setlength{\itemsep}{0pt}

    \item We introduce point supervision to TAL in wearable sensor
    streams to reduce temporal annotation costs.

    \item We propose PSEE to recover point-supervised pseudo boundaries by combining adaptive temporal ownership with semantic and sensor-transition cues.

    \item We conduct cross-subject experiments on four inertial-sensing
    benchmarks, demonstrating improved pseudo-boundary quality,
    detector compatibility, and robustness to point sampling.
\end{enumerate}

\section{Method}
\label{sec:method}

Our framework consists of point-localizer training, PSEE-based pseudo-segment
generation, and TAL detector training (Fig.~\ref{fig:framework}). Given a
sensor sequence $\mathbf{X}\in\mathbb{R}^{T\times D}$, supervision is provided
by point annotations $\mathcal{P}=\{(t_i,y_i)\}_{i=1}^{N}$, where $t_i$ lies
within action instance $i$ and $y_i$ denotes its class. PSEE recovers a pseudo
segment $(\tilde{s}_i,\tilde{e}_i,y_i)$ for each instance without ground-truth
boundaries, which then serves as supervision for standard TAL detectors.

\subsection{Point Localizer}
\label{subsec:localizer}

A lightweight temporal encoder $E$ with strided 1D
convolutions and dilated residual blocks extracts
$\mathbf{H}=E(\mathbf{X})\in\mathbb{R}^{T'\times d}$,
where $T'$ and $d$ denote the downsampled temporal length
and feature dimension. Two heads produce class logits
$Z_k^a(t)$, $k=1,\ldots,K$, and transition logits $Z^b(t)$.

We compute
$A_k(t)=\operatorname{AvgPool}_{5}(\sigma(Z_k^a(t)))$ for semantic localization, where $\sigma$ is the sigmoid function and
$\operatorname{AvgPool}_{5}$ averages over five temporal
steps. At annotated timestamps, training uses multi-class
cross-entropy on the class logits and an additional
positive activation term for the labeled class.

To supervise $B(t)=\sigma(Z^b(t))$ without boundary
annotations, we derive a sensor-specific prior from inputs. Let $\mathbf{q}_t\in\mathbb{R}^{M}$
concatenate the within-step channel-wise mean and
standard-deviation statistics. Define
$\boldsymbol{\Delta}_t=|\mathbf{q}_t-\mathbf{q}_{t-1}|$
and $\tau_t=\operatorname{median}_{m}(\Delta_{t,m})$.
The mean change magnitude and median-exceedance fraction are
\begin{equation}
\begin{aligned}
V(t)&=\frac{1}{M}\sum_{m=1}^{M}\Delta_{t,m},\\
C(t)&=\frac{1}{M}\sum_{m=1}^{M}
\mathbb{I}[\Delta_{t,m}>\tau_t],
\end{aligned}
\end{equation}
where $\mathbb{I}[\cdot]$ is the indicator function.
The unnormalized change score
$G(t)=V(t)[0.5+C(t)]$ weights the change magnitude by
median exceedance, with a nonzero baseline weight of $0.5$.
We normalize it as
\begin{equation}
P(t)=\operatorname{clip}\!\left(
\frac{G(t)-Q_{0.1}}{Q_{0.95}-Q_{0.1}+\epsilon},0,1\right),
\end{equation}
where $Q_{0.1}$ and $Q_{0.95}$ are the 10th and 95th
percentiles of $G(t)$;
$\operatorname{clip}(\cdot,0,1)$ restricts values to $[0,1]$;
and $\epsilon>0$ ensures numerical stability.

After aligning $P(t)$ to the localizer resolution by
linear interpolation, we train $B(t)$ using binary
cross-entropy against $P(t)$. Pseudo-boundary Inference stage uses $A_{y_i}(t)$ and the fused response
$\hat{B}(t)=[B(t)+P(t)]/2$ as semantic and transition
evidence respectively.

\subsection{Pseudo-boundary Inference (PBI) }
\label{subsec:psee}

PBI uses
$A_{y_i}(t)$ to estimate the extent of the labeled action and
$\hat{B}(t)$ to identify potential temporal separations.
It recovers one pseudo segment per annotated point by
restricting the search region, generating boundary candidates,
and jointly ranking candidate segments.

\textbf{Adaptive ownership.}
For ordered points $t_1<\cdots<t_N$, let $b_i$ denote
the location of the strongest $\hat{B}(t)$ response
between $t_i$ and $t_{i+1}$, using their midpoint when
the points are too close for reliable peak selection.
With $b_0$ and $b_N$ denoting the sequence start and
exclusive end, respectively, we define the ownership
region as $\Omega_i=[L_i,U_i)=[b_{i-1},b_i)$.
These limits constrain each instance's search range
without fixing its final action boundaries.

\textbf{Boundary candidate generation.}
Within $\Omega_i$, local Otsu thresholding of $A_{y_i}(t)$
followed by contiguous expansion from $t_i$ yields a
support interval $[c_i^L,c_i^R)$.
Let $p_i^L$ and $p_i^R$ locate the strongest $\hat{B}(t)$
responses on either side of $t_i$.
We combine support endpoints, transition peaks, and
ownership limits:
\begin{equation}
\mathcal{B}_i^L=\{L_i,c_i^L,p_i^L\},\quad
\mathcal{B}_i^R=\{U_i,c_i^R,p_i^R+1\},
\end{equation}
where $+1$ ensures an exclusive end.
After clipping, deduplication, and sorting, we pair the
endpoints, retaining only segments satisfying
$L_i\leq\ell\leq t_i<r\leq U_i$.

\textbf{Rank-based selection.}
Each candidate $[\ell,r)$ is evaluated by four criteria:
mean class activation inside the segment;
inside--outside mean activation contrast within $\Omega_i$;
mean boundary response
$[\hat{B}(\ell)+\hat{B}(r-1)]/2$;
and the fraction of $[c_i^L,c_i^R)$ covered.
We rank-normalize each criterion to $[0,1]$ across valid
candidates, with higher values indicating better scores,
and compute
\begin{equation}
S(\ell,r)=\frac{1}{4}\sum_{n=1}^{4}
\operatorname{Rank}_n(\ell,r).
\end{equation}
The highest-scoring candidate defines
$(\tilde{s}_i,\tilde{e}_i,y_i)$, with ties broken by
activation contrast and transition alignment.

\textbf{Segment reliability.}
We combine the selected score
$s_i=\operatorname{clip}(S(\tilde{s}_i,\tilde{e}_i),0,1)$
with boundary agreement $a_i$:
\begin{equation}
\begin{aligned}
a_i &= \operatorname{clip}\!\left(
1-\frac{\operatorname{IQR}(\mathcal{B}_i^L)+
\operatorname{IQR}(\mathcal{B}_i^R)}
{U_i-L_i},0,1\right),\\
\rho_i &= \operatorname{clip}\!\left(
\frac{2s_i a_i}{s_i+a_i+\epsilon},\epsilon,1\right),
\end{aligned}
\end{equation}
where $\operatorname{IQR}$ denotes the interquartile range.
This favors high-scoring segments with consistent
boundary estimates.

\subsection{TAL Training with Pseudo Boundaries}
\label{subsec:tal}

The pseudo segments and reliability scores from PBI form the
training set
$\widetilde{\mathcal{D}}
=\{(\tilde{s}_i,\tilde{e}_i,y_i,\rho_i)\}_{i=1}^{N}$.
After conversion to each detector's annotation format,
the pseudo segments replace ground-truth segments as training
targets. We retain the detector architecture and detector-specific
loss definitions, while using $\rho_i$ to reduce the influence
of uncertain pseudo boundaries on localization:
\begin{equation}
\mathcal{L}_{\mathrm{TAL}}
=
\mathcal{L}_{\mathrm{cls}}
+
\lambda_{\mathrm{loc}}
\frac{\sum_i \rho_i \mathcal{L}_{\mathrm{loc}}^{(i)}}
     {\sum_i \rho_i},
\end{equation}
where $\mathcal{L}_{\mathrm{cls}}$ is the detector's classification
loss, $\mathcal{L}_{\mathrm{loc}}^{(i)}$ is its localization loss
associated with pseudo instance $i$, and $\lambda_{\mathrm{loc}}$
balances the two terms. Reliability weighting applies only to
localization; the classification loss retains its original
formulation and is not reweighted by $\rho_i$.

\section{Experiments}
\label{sec:experiment}

\subsection{Experimental Setup}
\label{subsec:setup}

We evaluate PSEE on four inertial-sensing benchmarks:
WEAR~\cite{bock2024wear}, WetLab~\cite{scholl2015wearables},
SBHAR~\cite{reyesortiz2016transition}, and
RWHAR~\cite{sztyler2016body}, all under leave-one-subject-out (LOSO)
evaluation. Since native point annotations are unavailable, one point per
action instance is sampled from a Gaussian centered at the segment midpoint
with standard deviation equal to one sixth of the segment length. All methods
use the same sampled points, random seed, and data splits. TAL performance is
reported as mAP over tIoU thresholds $0.3{:}0.1{:}0.7$. We additionally
report pseudo-boundary average tIoU against ground-truth segments to evaluate
boundary quality independently of the downstream detector.

For controlled comparison, SF-Net~\cite{ma2020sf} and
OPOT~\cite{xia2024timestamp} are adapted to generate pseudo segments from the
same point annotations. SFNet-Adapt extracts contiguous action regions from
point-guided responses, while OPOT-Adapt converts its dense OT labels into
the action region containing each annotated point. All generated pseudo
segments are used to train the same downstream detector,
SlimSTAD~\cite{cui2026light}, adapted to 1D sensor features.

\begin{table*}[t]
\centering
\caption{Comparison with point-supervised baselines using the same
SlimSTAD~\cite{cui2026light} detector.}
\label{tab:baseline_comparison}

\footnotesize
\setlength{\tabcolsep}{3.2pt}
\renewcommand{\arraystretch}{1.12}

\begin{tabular}{lcccccccccccccccc}
\toprule
& \multicolumn{4}{c}{WEAR~\cite{bock2024wear}}
& \multicolumn{4}{c}{WetLab~\cite{scholl2015wearables}}
& \multicolumn{4}{c}{RWHAR~\cite{sztyler2016body}}
& \multicolumn{4}{c}{SBHAR~\cite{reyesortiz2016transition}} \\
\cmidrule(lr){2-5}
\cmidrule(lr){6-9}
\cmidrule(lr){10-13}
\cmidrule(lr){14-17}

Method
& @0.3 & @0.5 & @0.7 & mAP
& @0.3 & @0.5 & @0.7 & mAP
& @0.3 & @0.5 & @0.7 & mAP
& @0.3 & @0.5 & @0.7 & mAP \\
\midrule

OPOT-Adapt~\cite{xia2024timestamp}
& 82.77 & 72.28 & 55.81 & 70.68
& 35.76 & 22.49 & 8.20  & 21.67
& 12.34 & 9.32  & 5.20  & 8.98
& 68.79 & 42.73 & 17.36 & 42.48 \\

SFNet-Adapt~\cite{ma2020sf}
& \textbf{84.85} & \textbf{76.15} & 57.92 & \textbf{74.32}
& 18.94 & 6.85  & 1.97  & 9.12
& 6.28  & 1.93  & 1.67  & 3.04
& \textbf{85.99} & 47.01 & 14.57 & 50.23 \\

\textbf{PSEE}
& 84.81 & 75.76 & \textbf{58.83} & 74.14
& \textbf{46.43} & \textbf{33.44} & \textbf{19.97} & \textbf{33.55}
& \textbf{76.82} & \textbf{67.39} & \textbf{47.73} & \textbf{64.68}
& 61.64 & \textbf{52.70} & \textbf{35.17} & \textbf{50.44} \\

\bottomrule
\end{tabular}
\vspace{-7pt}
\end{table*}

\vspace{-8pt}
\subsection{Comparison with Point-supervised Baselines}
\label{subsec:comparison}

We compare PSEE with SFNet-Adapt~\cite{ma2020sf} and
OPOT-Adapt~\cite{xia2024timestamp} using the same
SlimSTAD~\cite{cui2026light} detector, so the differences mainly reflect the
quality of point-derived temporal supervision. As shown in
Table~\ref{tab:baseline_comparison}, PSEE brings particularly large gains on
WetLab and RWHAR, achieving 33.55\% and 64.68\% mAP compared with 21.67\%
and 8.98\% for the stronger baseline on each dataset.

More importantly, the advantage becomes clearer under stricter temporal
overlap. PSEE achieves the highest at tIoU $0.7$ on all four datasets.
On SBHAR, although SFNet-Adapt is substantially better at tIoU $0.3$
(85.99\% vs.\ 61.64\%), PSEE surpasses it at $0.5$ and $0.7$, reaching
52.70\% and 35.17\%. A similar advantage appears on WEAR at tIoU $0.7$
(58.83\% vs.\ 57.92\%). This pattern suggests that PSEE improves boundary
precision rather than merely coarse action coverage, consistent with its
goal of recovering point-supervised temporal extents from sparse points.

\vspace{-8pt}
\subsection{Generality across TAL Detectors}
\label{subsec:detector_generality}

We further evaluate whether PSEE supervision transfers across different TAL
detectors. As shown in Table~\ref{tab:detector_generality}, the same
PSEE-generated pseudo segments are used to train SlimSTAD,
ActionFormer, and TriDet. All three detectors achieve competitive results
under this shared supervision, with mAP ranging from 50.44\% to 81.19\% on
SBHAR and from 20.45\% to 64.68\% on RWHAR. The substantial variation across
detectors indicates that PSEE is not tied to a specific TAL architecture,
while final performance remains influenced by the downstream detector. This motivates evaluating pseudo-boundary quality separately from mAP.

\vspace{-8pt}
\subsection{Pseudo-boundary Quality}
\label{subsec:pseudo_quality}

To evaluate pseudo-boundary quality independently of downstream detector
optimization, we measure average tIoU between generated pseudo segments and their
ground-truth segments, providing a direct assessment of temporal extent
recovery from sparse point annotations.

\begin{figure}[t]
    \centering
    \includegraphics[width=0.9\columnwidth]{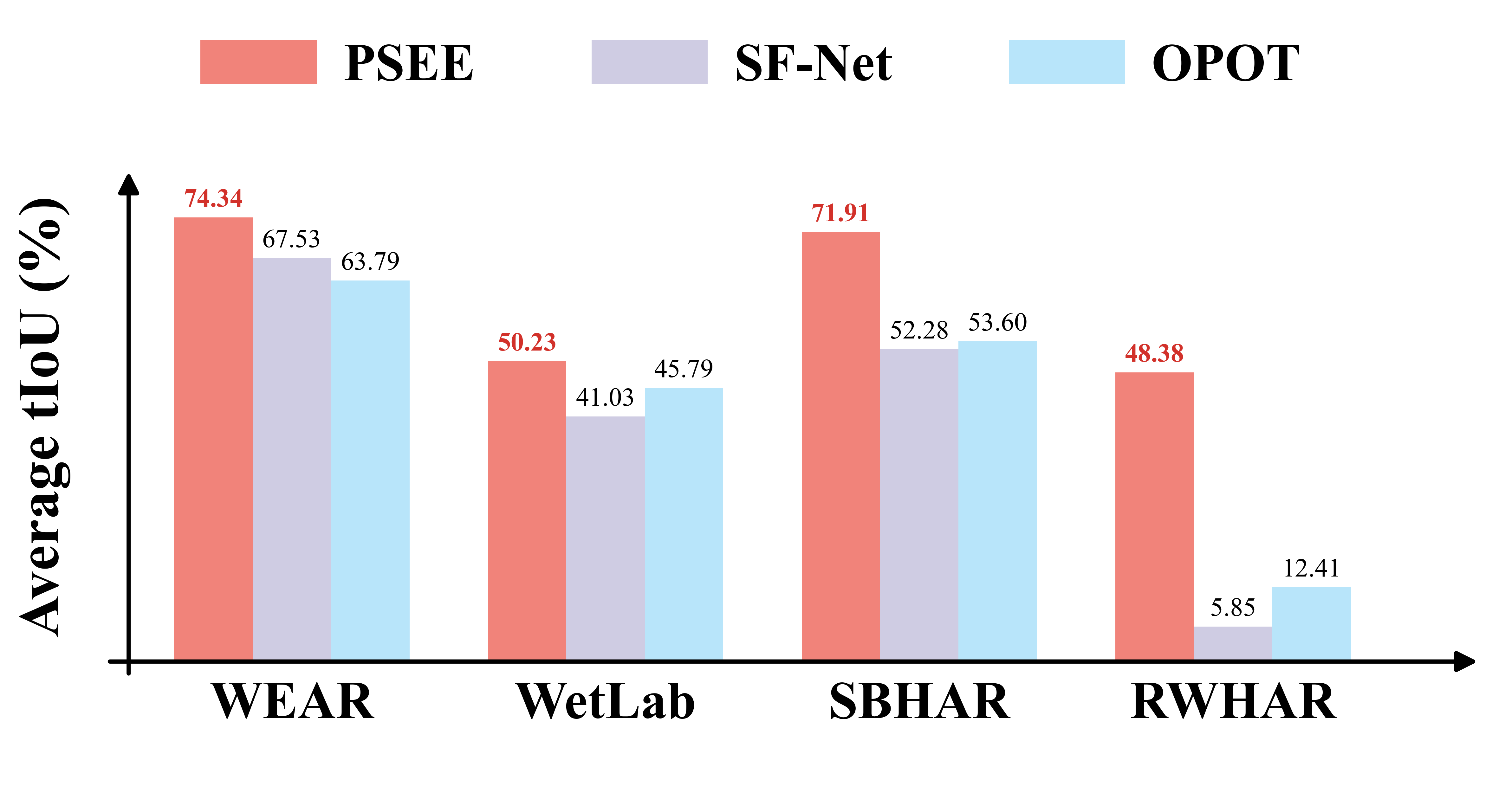}
    \caption{Pseudo-boundary quality on four sensor datasets, measured by
    average tIoU between generated pseudo segments and ground-truth
    boundaries.}
    \label{fig:pseudo_analysis}
    \vspace{-5pt}
\end{figure}

As shown in Fig.~\ref{fig:pseudo_analysis}, PSEE achieves 74.34\%, 50.23\%,
71.91\%, and 48.38\% average tIoU on WEAR, WetLab, SBHAR, and RWHAR, respectively,
outperforming both adapted baselines on four datasets. The gains over the
stronger baseline are especially large on SBHAR and RWHAR, reaching 18.31 and
35.97 percentage points. This demonstrates that PSEE more accurately expands
sparse points into complete point-supervised temporal extents, with the improvement
arising before downstream TAL detector training.

\begin{table}[t]
\centering
\caption{Detector generality of PSEE. Results are mAP over tIoU thresholds $0.3{:}0.1{:}0.7$.}
\label{tab:detector_generality}

\footnotesize
\setlength{\tabcolsep}{1.7pt}
\renewcommand{\arraystretch}{1.12}

\begin{tabular}{lcccc}
\toprule
Detector
& WEAR
& WetLab
& RWHAR
& SBHAR
 \\
\midrule

SlimSTAD~\cite{cui2026light}
& \textbf{74.14}
& \textbf{33.55}
& \textbf{64.68}
& 50.44
 \\

ActionFormer~\cite{zhang2022actionformer}
& 66.66
& 28.68
& 20.45
& 80.62
 \\

TriDet~\cite{shi2023tridet}
& 66.42
& 28.21
& 33.31
& \textbf{81.19}
 \\

\bottomrule
\end{tabular}
\vspace{-5pt}
\end{table}

\begin{table}[t]
\centering
\caption{Comparison of point sampling strategies for PSEE.
Results are reported as average tIoU (\%).}
\label{tab:point_sampling}

\footnotesize
\setlength{\tabcolsep}{2.8pt}
\renewcommand{\arraystretch}{1.12}

\begin{tabular}{lcccc}
\toprule
Sampling
& WEAR
& WetLab
& RWHAR
& SBHAR \\
\midrule

Uniform
& 72.79
& 49.25
& 40.98
& 69.43 \\

Gaussian
& \textbf{74.34}
& \textbf{50.23}
& \textbf{48.38}
& \textbf{71.91} \\

\bottomrule
\end{tabular}
    \vspace{-5pt}

\end{table}

\begin{table}[t]
\centering
\caption{Component ablation of PSEE under Gaussian point sampling.
Results are average tIoU (\%); $\checkmark$ and $\times$ denote enabled and
disabled components.}
\label{tab:component_ablation}

\footnotesize
\setlength{\tabcolsep}{2.2pt}
\renewcommand{\arraystretch}{1.12}

\begin{tabular}{cccccc}
\toprule
Trans. & Own.
& WEAR
& WetLab
& RWHAR
& SBHAR \\
\midrule

$\checkmark$ & $\times$
& \textbf{74.71}
& 48.37
& {39.53}
& {59.88} \\

$\times$ & $\times$
& 71.03
& {49.82}
& 39.36
& 57.68 \\

$\checkmark$ & $\checkmark$
& {74.34}
& \textbf{50.23}
& \textbf{48.38}
& \textbf{71.91} \\

\bottomrule
\end{tabular}
    \vspace{-5pt}

\end{table}

\subsection{Ablation and Robustness Analysis}
\label{subsec:ablation}

\textbf{Robustness to point sampling.}
Table~\ref{tab:point_sampling} compares the Gaussian sampling used in the
main experiments with uniform sampling over each action interval, using
pseudo-boundary average tIoU as the evaluation metric. Under uniform sampling, PSEE
retains comparable average tIoU on WEAR, WetLab, and SBHAR (72.79\%, 49.25\%, and
69.43\%), with a larger drop on RWHAR (40.98\% vs.\ 48.38\%). These results
show that PSEE does not critically depend on points being concentrated near
action centers and remains effective under broader point distributions.

\textbf{Component ablation.}Table~\ref{tab:component_ablation} examines transition cues and adaptive
ownership. Since adaptive ownership determines inter-point separators from
transition responses, removing transition cues also disables adaptive
ownership. The transition-only variant retains transition-based boundary
candidates and scoring, but replaces adaptive separators with neighboring-point
midpoints.

Adding transition cues to midpoint ownership improves avg tIoU on WEAR, RWHAR,
and SBHAR, while the gain is limited on WetLab. Further enabling adaptive
ownership substantially improves RWHAR from 39.53\% to 48.38\% and SBHAR
from 59.88\% to 71.91\%, with a smaller gain on WetLab and comparable
performance on WEAR. These results show that transition-guided adaptive
ownership is particularly important for accurate pseudo-boundary recovery
on RWHAR and SBHAR.

\section{Conclusion}

We presented PSEE, which expands sparse action points into
point-supervised pseudo segments for wearable-sensor temporal action
 localization. Experiments on four inertial-sensing
benchmarks demonstrate improved pseudo-boundary quality
over adapted point-supervised baselines and compatibility
with different TAL detectors. Component ablations highlight
the value of combining transition cues with adaptive
ownership for boundary recovery. 
By reducing reliance on precise boundary annotations, PSEE
advances annotation-efficient activity localization in wearable
sensing and offers a path toward scalable point-supervised activity
understanding.

\bibliographystyle{IEEEbib}
\bibliography{strings,refs}

\end{document}